# Role of Dependency Distance in Text Simplification: A Human vs ChatGPT Simplification Comparison

Sumi Lee, MA[1], Gondy Leroy, PhD[2], David Kauchak, PhD[3], Melissa Just, PhD[4]
[1,2] University of Arizona, AZ; [3]Pomona College, CA; [4]University of Saskatchewan, Canada

**Abstract**
*This study investigates human and ChatGPT text simplification and its relationship to dependency distance. A set of 220 sentences, with increasing grammatical difficulty as measured in a prior user study, were simplified by a human expert and using ChatGPT. We found that the three sentence sets all differed in mean dependency distances: the highest in the original sentence set, followed by ChatGPT simplified sentences, and the human simplified sentences showed the lowest mean dependency distance.*

**Introduction**
Enhancing the understandability of biomedical information is vital in fostering health-literate patients. However, empirical evidence shows that readability formulas are not appropriate tools [1], [2]. Systematic examination of various textual descriptive statistics and syntactic complexity can contribute to creating more accessible and readable biomedical texts. Dependency distance is a metric of syntactic complexity that calculates the syntactic tree nodes in a sentence. This study evaluates dependency distance as a metric for text simplification.

**Methods**
A total of 220 sentences with increasing grammatical difficulty were used [3]. The grammatical difficulty was calculated by parsing the 3rd level in the parse tree of all sentences in English Wikipedia and calculating the frequency of each structure. These frequencies correlate with syntactic complexity. A random selection of 220 sentences from 11 different grammar frequency bins were simplified by a medical librarian by only changing the grammatical structure and also by ChatGPT-3.5 (OpenAI) with the prompt "Simplify these sentences without changing any of the words, but just the syntax". The dependency distance of these three sets was calculated by a python package (TextDescriptives, spaCy v.3 pipeline components).

**Results**
Overall, the mean dependency distance of original sentences in the different grammatical bins did not differ (Figure 1). Comparing the method of simplification showed that the mean dependency distances of the original sentence set (M=2.90, SD=0.63) decreased when simplified using human or ChatGPT simplification (Figure 2). However, ChatGPT simplified sentences (M=2.87, SD=0.62) showed a slightly higher dependency distance than human simplified sentences (M=2.77, SD=0.60). A one-way ANOVA with mean dependency distances as the dependent variable and sentence types (original, ChatGPT simplified, human simplified) as the independent variable showed that there was a trend but no significant effect of sentence type (F(2,657)=2.62, p=0.07).

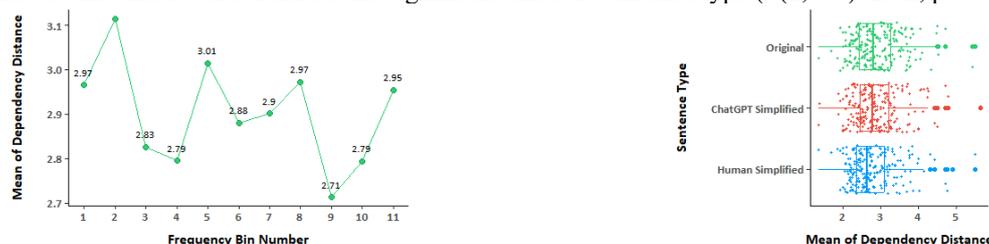

**Figure 1.** Mean dependency distance of three sentence types (Left). Mean dependency distance of original sentences in different frequency bins (Right),

**Acknowledgements**
Research reported in this paper was supported by the U.S. National Library of Medicine of the National Institutes of Health under Award Number R01LM011975. The content is solely the responsibility of the authors and does not necessarily represent the official views of the National Institutes of Health.